\documentclass[letterpaper, 10 pt, conference]{ieeeconf}

\IEEEoverridecommandlockouts                              	
\usepackage{booktabs}
\usepackage{cite}
\usepackage{amsmath,amssymb,amsfonts}
\usepackage{graphicx}
\usepackage[caption=false]{subfig}
\usepackage{algpseudocode}
\usepackage{textcomp}
\usepackage{hyperref}
\usepackage{setspace}  

\newcounter{algorithm}

\usepackage[normalem]{ulem}
\usepackage{xcolor}

\newcommand\subs[1]{_{\text{#1}}}

\algnewcommand{\LeftComment}[1]{\Statex \(\triangleright\) #1}

\def\BibTeX{{\rm B\kern-.05em{\sc i\kern-.025em b}\kern-.08em
    T\kern-.1667em\lower.7ex\hbox{E}\kern-.125em}}

\usepackage{aveasmargins}
\begin{document}

\title{\LARGE \bf
Divide and Conquer: A Systematic Approach for Industrial Scale High-Definition OpenDRIVE Generation from Sparse Point Clouds
}

\author{Leon Eisemann$^{1}$ and Johannes Maucher$^{2}$
\thanks{*This publication was written in the context of the AVEAS research project (www.aveas.org), funded by the German Federal Ministry for Economic Affairs and Climate Action (BMWK) within the program ``New Vehicle and System Technologies''.}
\thanks{$^{1}$Leon Eisemann is with the department of Artificial Intelligence \& Big Data at Porsche Engineering Group GmbH, 71287 Weissach, Germany}%
\thanks{$^{2}$Johannes Maucher is with the Institute for Applied Artificial Intelligence at
Stuttgart Media University, 70569 Stuttgart, Germany}%
}

\maketitle
\thispagestyle{empty}
\pagestyle{empty}

\aveasSetMargins{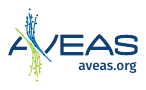}
\aveasSetIEEEFoot{2024}
\aveasSetIEEEHead{L. Eisemann and J. Maucher, ``Divide and Conquer: A Systematic Approach for Industrial Scale High-Definition OpenDRIVE Generation from Sparse Point Clouds,'' 2024 IEEE Intelligent Vehicles Symposium (IV), Jeju Island, Republic of Korea, 2024, pp. 2443-2450}{10.1109/IV55156.2024.10588602}

\begin{abstract}
High-definition road maps play a crucial role in the functionality and verification of highly automated driving functions. These contain precise information about the road network, geometry, condition, as well as traffic signs. 
Despite their importance for the development and evaluation of driving functions, the generation of high-definition maps is still an ongoing research topic. 
While previous work in this area has primarily focused on the accuracy of road geometry, we present a novel approach for automated large-scale map generation for use in industrial applications.
Our proposed method leverages a minimal number of external information about the road to process LiDAR data in segments. These segments are subsequently combined, enabling a flexible and scalable process that achieves high-definition accuracy. Additionally, we showcase the use of the resulting OpenDRIVE in driving function simulation.
\end{abstract}

\setstretch{0.99}  
\section{Introduction} \label{introduction}
The introduction of Advanced Driver Assistance Systems (ADAS) has brought a significant shift in the automotive industry, particularly in the development of automated driving (AD) functions that achieve SAE Level 3 and above. These systems have become increasingly complex, by combining the hardware / software systems capable of operating under these conditions, as well as an expanded operational design domain (ODD).
Further, the functional safety and the safety of the intended functionality (SOTIF) must be ensured across a growing number of traffic situations.
These shifts imply the need for higher requirements on test and evaluation of these functions, where more and more test kilometers are needed. Based on the average driven kilometers between two fatal accidents on German highways, Wachenfeld et al. \cite{wachenfeld2016} estimate this number to be 6.61 billion kilometers.
With these prerequisites, real-world test drives are no longer feasible, and the adoption of simulation-based safety evaluation is critical for the success of AD functions. As a consequence also the realism and accuracy of simulations have to be enhanced and derived from the real world. 
High-definition (HD) maps of the road are considered an integral part of every simulation, aside from their use in AD functions\cite{xinxin-csg-2020}. 
Albeit their importance in functions and simulations, their creation, especially on a large scale, is still an open research topic.
To model roads as a basis for AD simulation, the approaches can be divided into four different categories\cite{bao-high-definition-2022}:
\begin{figure}[t]
\centering
\includegraphics[width=1\columnwidth]{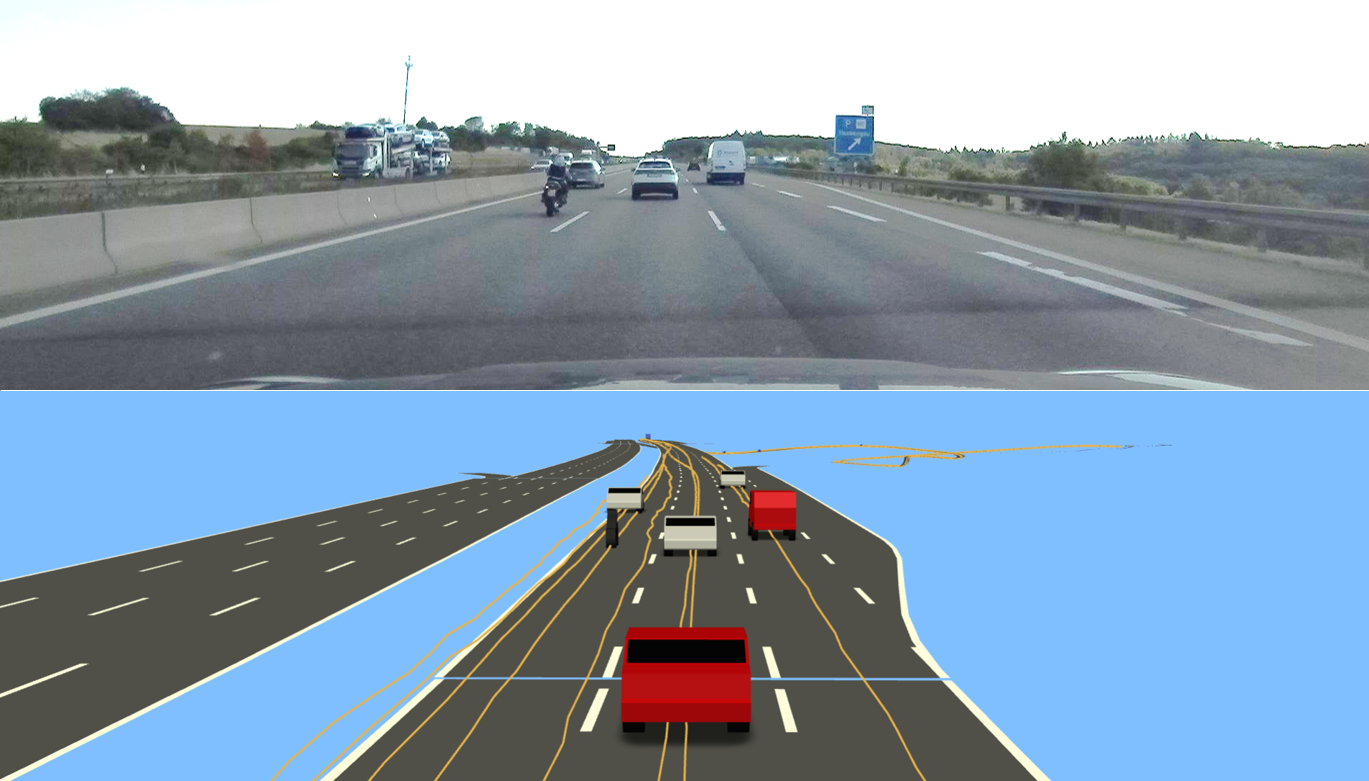}
\caption{Representation of the output of our proposal. On the top the camera view of the ego vehicle. At the bottom, the resulting OpenDRIVE and OpenScenario files of our method, displayed in esmini \cite{esmini-environment-2023}.}
\label{fig:real-to-esmini-comparison}
\end{figure}

First, the simple conversion of different standard-definition (SD) map sources or aerial images into the respective format needed for simulation. 

Second, the generation of HD maps from mobile mapping platforms. Mobile mapping platforms hereby describe vehicles fitted with a high number of sensors to collect detailed environmental information. Commonly, these platforms employ a combination of Global Navigation Satellite System (GNSS), Inertial Measurement Unit (IMU), and Light Detection and Ranging (LiDAR) sensors.

Third, the fusion of SD information with mobile mapping to reduce shortcomings of the individual methodologies\cite{bao-high-definition-2022}.

Lastly, especially in automotive simulation, HD maps are often created by specialized service providers, for example in \cite{montanari-maneuver-based-2021, richter-systematische-2016, schwab-spatio-semantic-2020}. 
However, given that these providers operate commercially, their methods and the level of automation are not publicly available. Consequently, they are not examined within the scope of this work.

In the field of automotive simulation, the choice of a map source is use case dependent\cite{aveas-paper-2023}. As described by Eisemann et al. \cite{aveas-paper-2023}, for a comprehensive traffic scenario extraction, trajectory and map data need to be referenced to one another, ideally created jointly to minimize measurement offsets. This focuses on the use of mobile mapping data since these platforms allow a flexible acquisition. 

Nevertheless, research on HD map generation from mobile mapping platforms is limited. Existing studies often depend on specific sensor setups and mounting positions\cite{eisemann-opendrive}. This results in high costs for equipment, sensors, and workforce for the setup and maintenance of these platforms. Further, the presented methods are focused on the accuracy of the results rather than the large-scale acquisition.
Considering that simulations should provide a comprehensive evaluation of the function under test, it is crucial for maps to offer extensive coverage and capture the variety of the world. This highlights the need for further research in the field of large-scale map generation.

In this paper, we therefore propose a generic approach for HD map generation incorporating the aforementioned aspects. One exemplary scene is presented in Fig.\ref{fig:real-to-esmini-comparison}.

We aim to provide a scalable approach, capable of handling multiple hundred kilometers of drive data, with minimal reliance on external information.
Further, our methodology, described in Sec.~\ref{sec:method}, focuses on utilizing sparse LiDAR data obtained from research vehicles, enabling a more cost-effective acquisition process.
In comparison, the current state-of-the-art approaches are constrained by specific sensor setups and have only reached several hundred meters in length, as presented in Sec.~\ref{sec:background}.
In Sec.\ref{sec:results-discussion}, we demonstrate that our proposed approach achieves HD map accuracy and we showcase the quality through a trajectory-based simulation derived from the original test drive.

\section{Background}\label{sec:background}
The following section reviews existing mapping approaches and introduces the map standard used in this work.

\subsection{Related Work}\label{subsec:related-work}
Current research in the generation of HD maps is primarily concerned with developing methods for mapping the local surroundings of the vehicle\cite{dong2022SuperFusion}. 
These extensively studied the use of machine learning based bird's eye view sensor fusion \cite{li2021hdmapnet}, road network graph detection by transformer networks \cite{xu2022rngdet++}, or hierarchical query embedding transformers \cite{MapTR}. 
Albeit their impressive results, these methods introduce multiple limitations for industrial applications. Since machine learning models build the basis of these methods, they also introduce associated requirements, like available training data, the resulting domain gap, and challenges such as catastrophic forgetting \cite{Kalb_2023_CVPR}. 
Moreover, the results presented by these methods are often in the form of plots or vectors, rather than a standardized map format\cite{dong2022SuperFusion, li2021hdmapnet, xu2022rngdet++, MapTR}. Since the map format has a significant impact on the processing, these methods are not further examined.

Research on the export of a standardized HD map format is relatively sparse\cite{bao-high-definition-2022}. Chiang et al. \cite{chiang_automated_2022} present an approach for exporting OpenDRIVE, by using the trajectory of the ego vehicle with an installed ground-facing laser scanner. Using extracted road edges, lane marking classification, and trajectory information the authors employ a multistage reconstruction process for the reconstruction of road geometry.

A specialized approach for the extraction of traffic scenarios is shown by Karunakaran et al.\cite{karunakaran-automatic-2022}. The data used in this study was obtained from a vehicle equipped with six LiDAR sensors and an additional ground-facing LiDAR. By taking advantage of the line-wise LiDAR scanning pattern, the authors filter out lane markings which are transformed into Lanelet2 format and subsequently converted to OpenDRIVE.

Focusing on the independence from external data and the use of sparse LiDAR data, we introduced a method for the automated OpenDRIVE and OpenSCENARIO generation in \cite{eisemann-opendrive}. 
The algorithm involves reflectivity-based lane marking extraction, 3D lane line generation, and estimation of the reference line without relying on the prior road information.

Although these methods achieve a standardized map format, the reliance on special sensor characteristics \cite{karunakaran-automatic-2022} or mounting positions \cite{chiang_automated_2022} hinder the adaption of these methodologies. 
Moreover, generating OpenDRIVE geometry as one consecutive road, as presented in our previous work\cite{eisemann-opendrive}, can lead to errors in large-scale road reconstruction. 

Our motivation is to achieve a precise road representation, achieving HD map quality, without a vast number of sensors, multi-sensor fusion, or machine learning. In addition, we aim for a scalable process, capable of handling multiple hundred kilometers of road network. Therefore, we focus on the parallel execution of process steps, the limited impact of reconstruction errors on the overall map, and the possibility of manual intervention.

Based on the aforementioned requirements, we base our feature extraction on our previous research presented in \cite{eisemann-opendrive}. This is due to the fact, that the approach has been demonstrated to achieve stable HD map quality while utilizing only sparse 72° field of view (FoV) LiDAR information.

\subsection{Map Format}\label{sec:mapformat}
We select ASAM OpenDRIVE \cite{asam-opendrive} as our targeted HD map format. One of the main reasons is the wide-scale adoption in the automotive industry and the integration into tools and software platforms for virtual testing\cite{chiang_automated_2022, aveas-paper-2023}. 

OpenDRIVE describes the overall course of the road along a reference line. Extended elements such as elevation, super-elevation, and lanes are attached along this line. Inspired by road planning and construction, a reference line can consist of different geometric elements, namely \textit{line}, \textit{spiral}, \textit{arc}, or \textit{paramPoly3} (parametric cubic curves). One or more of these geometries can be used or concatenated to mathematically describe the detailed road structure.

Another critical element for the description of road networks is the relation between individual roads and lanes. In OpenDRIVE, entry and exits of roads are described as individual junctions, where the predecessor/successor relation between each lane is defined.

\begin{figure*}[t]
\centering
\includegraphics[width=2\columnwidth]{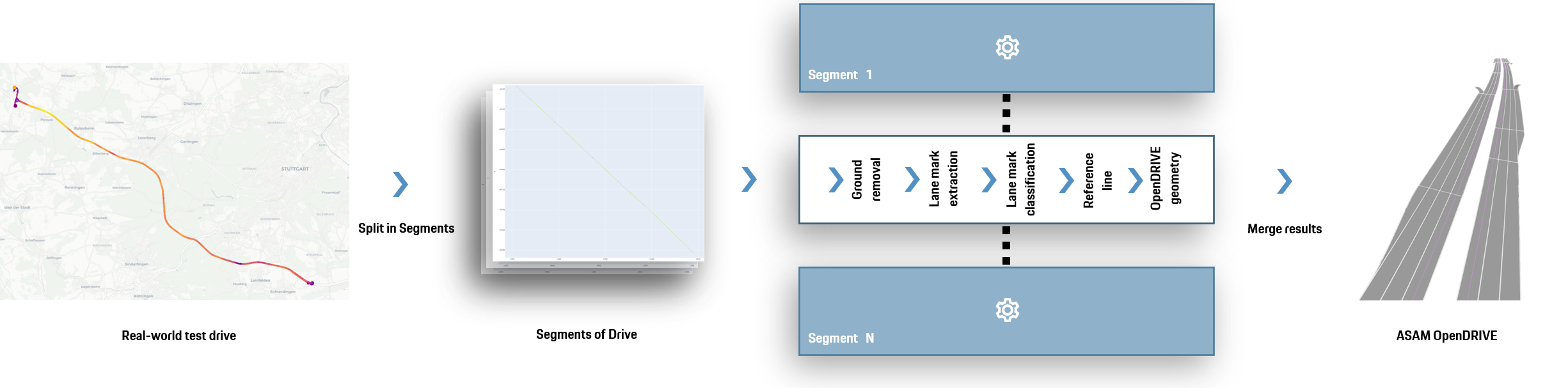}
\caption{General overview of our method. Real-world test drive data is split into segments, using OpenStreetMap information. 
 LiDAR data is accumulated and processed within each segment.
By extracting lane markings and combining them with government regulations and OpenStreetMap data, the reference line is estimated. 
The resulting geometric elements are compiled into a single OpenDRIVE file, where lane relations and junctions are also defined.}
\label{fig:process-overview}

\end{figure*}

\section{Method}\label{sec:method}

As shown in Fig.~\ref{fig:process-overview}, our proposal consists of multiple processing steps, which are further described in the subsequent sections. Since the core principle of our approach is the division of LiDAR data into segments and fusion with external information, Sec.~\ref{subsec:data-and-processing} describes the initial data used and the segmentation process. The subsequent sections describe the processing steps performed on each separate segment, which are then combined into the final OpenDRIVE in Sec.~\ref{subsec:export-opendrive}.

\subsection{Data \& Preprocessing}\label{subsec:data-and-processing}
Similar to other approaches, we also adapt LiDAR data for the extraction of accurate road information. 
With the decreasing costs for LiDAR sensors over the last few years, their integration into both research and production vehicles has become more common.
LiDAR sensors provide high positional accuracy of detections, while the measured reflectivity allows easier detection of road markings.

For collecting accurate environment and traffic behavior data, we make use of the JUPITER vehicle described by Haselberger et al. \cite{Haselberger-Jupiter}. The platform consists of a Porsche Cayenne fitted with multiple Livox LiDARs and GeneSys ADMA GNSS inertial system. To demonstrate the robustness of our approach, we solely use the front-mounted Livox LiDAR sensor, which possesses a complex and inhomogeneous scanning pattern. 
In contrast to previous work, the data collection was not focused on road generation but on the acquisition of critical driving scenarios as part of the AVEAS~\cite{aveas-paper-2023}~\footnote{\url{www.aveas.org}} research project.
In order to record authentic traffic behavior, data was collected mainly on the German interstate highways A8 and A81 around Stuttgart at peak traffic times. The speed profile of the recordings ranges from 80 to 140 km/h in the considered parts.

Throughout this work, we further leverage data from the OpenStreetMap~\footnote{\url{www.openstreetmap.org}} (OSM) project as an external data source. OSM is a crowd-sourced map platform that provides comprehensive information about road networks, connections, and surroundings. Since our proposed method requires minimal prior information for the processing, we only use the following data from the OSM project:
\begin{itemize}
    \item country of the recording
    \item type of road, e.g., highway, entry/exit
    \item number of lanes
    \item existence of shoulder lanes
    \item connection to exit or entry lanes
\end{itemize}
In parts of our segment generation process, we further utilize the road segment description of OSM to identify road sections with constant properties. This process is detailed in the following section. We note that the listed information is taken from OSM for ease of use and could also be derived from other sources such as image processing.

\subsection{Segment Generation}\label{subsec:segment-generation}
Dividing the overall road extraction into segments serves multiple purposes. 

First, enabling the parallel processing of multiple segments. Especially as the extraction and analysis of LiDAR data is a computationally intensive task, the distribution over multiple workers, for example in a cloud environment, can significantly reduce the overall generation time.

Second, in case of processing failures, only the specific segment is affected and can be either dismissed or corrected.

Lastly, as some road geometry corner cases can be labor-intensive to automate, a segment-based approach allows for easier manual generation or correction of such cases.

For splitting up a real-world test drive into segments we utilize the road geometry provided by OSM. 
OSM describes roads as \textit{ways}, which represent parts of a road with constant attributes, e.g., number of lanes, and shoulder lanes. Geometry and connection of a \textit{way} is defined by \textit{nodes}. The overall course of the road is outlined by \textit{edges}, which are implicitly defined through the connection of the \textit{nodes}. Extracting these \textit{edges} creates a representation of the road network usable in map matching. For the segmentation of our process, we choose the division into the individual \textit{edges}, since these represent the smallest building blocks of the OSM road network.

We make use of the map matching framework introduced by Meert et al. \cite{meert2018hmm} to assign each timestep to the corresponding \textit{edge}. In contrast to the authors, matching is done for each timestep individually, without leveraging a Hidden Markov Model (HMM). The absence of HMM aims to prevent failures in the case of insufficient observations resulting in ambiguous solutions.
We note, that our approach may yield inconsistent results, such as mismatches on overpasses, and is subject to further research.

For each \textit{edge}, the OSM information detailed in Sec.~\ref{subsec:data-and-processing} is stored, together with LiDAR and GNSS data matched to the specific \textit{edge}.
If a single recording drive passes an \textit{edge} multiple times, the respective timesteps are combined, if the time difference is less than eight hours, assuming that the course of the road is unchanged within one day. 

Through the use of a unified data loading architecture, the processing of the individual segments is independent of one another and can be parallelized, either over multiple cores or multiple cloud workers. 

The processing of each segment is highly inspired by our previous work in \cite{eisemann-opendrive}, which has shown high stability in line marking extraction with the use of sparse point clouds.

\subsection{Line Marking Extraction} \label{subsec:line-marker-extrac}
For each matched segment timestep, the corresponding LiDAR and GNSS measurements are synchronized. 

Following \cite{eisemann-opendrive}, we discard LiDAR points above the sensor mounting position to reduce point cloud size.  
In addition, we also crop the point cloud based on the maximum road width derived from OSM lane information. 
As the ego vehicle's overall position on the road is unknown at this stage, the possible maximum width is used on both sides of the ego vehicle. Despite this high threshold value, results on our data show a significant reduction of outliers, especially when the ego vehicle is on the outermost lane of the road.

To retrieve the corresponding points belonging to the road surface, we leverage a \textit{RANSAC}\cite{ransac-fischler}-based approach to estimate the ground plane. To account for sloping roadsides, all plane inliers, as well as points below the ground plane, are taken into consideration for further processing steps.

The resulting filtered point clouds are transformed into a unified segment coordinate system. Therefore, the global ego transformation is estimated based on GNSS coordinates and IMU rotation. Through a standardized transverse Mercator projection, GNSS coordinates are transformed into Euclidean space.
We define the first global transformation as the origin of the segment coordinate system, corresponding to the vehicle coordinate system of the first timestep.

The extraction of the individual lane markings exploits their higher reflectivity compared to the road surface. Depending on the condition or aging of the lane marking their reflection is decreased. Therefore, our process first makes use of a filter threshold of 25\% to determine LiDAR returns belonging to markings. If the subsequent processing is not able to build clusters or find a consecutive lane marking, the filtering is repeated with a 5\% lowered threshold.
Since reflectivity-based filtering leads to considerable outliers, an additional radius-based outlier reduction is performed\cite{eisemann-opendrive}.

\subsection{Generation of Lane Marking}\label{sec:gen-candid-lines}
The evaluation of continuous lane markings is based on the assumption that lane markings are arranged in a row along the edges of each lane\cite{frank-leitfaden-2017}. Further, the overall process assumes that lane markings, dashed as well as solid, are elongated in the direction of the lane and therefore in the direction of the next marking \cite{frank-leitfaden-2017}. As these considerations are true in most non-crossing situations, see \cite{frank-leitfaden-2017}, we widely adopt the lane generation process described in \cite{eisemann-opendrive}.

This consists of clustering the filtered lane marking points to associate LiDAR returns to individual markings, enabling the estimation of their direction on which consecutive markings are connected to build lane boundaries.

Given the unequal length and number of contained markings in each segment, DBSCAN \cite{dbscan-ester} clustering is used. 
The findings in \cite{eisemann-opendrive} and further analysis of our data imply, that solid lane markings can result in a single cluster over the full length of the segment.
These solid clusters contain accurate information about the course of the road, even in-between dashed markings. 
In addition, they provide reliable information about the ego position on the road.
To extract this information for further processing, markings stretched over a length of 12 m are split into 6 m parts, which equals the size of a single dashed marking\cite{frank-leitfaden-2017}. 
Since the consecutive steps in Sec.~\ref{subsec:comb-candid-lines} and Sec.~\ref{subsec:classification-lane-markings} depend on lane marking relations and classifications, clusters identified as solid get assigned the corresponding class and connection.

Since the reconstruction of lane marking relation in point clouds is a non-trivial task, related research, e.g., \cite{karunakaran-automatic-2022, chiang_automated_2022}, leverages the ego trajectory. While these methods provide stable results on consistent drives, they can yield mismatched relations in the case of ego vehicle lane changes. 

Utilizing the mentioned lane marking assumptions allows deriving the lane marking relation without depending on the ego trajectory.
Therefore, we leverage the elongation of the lane marking clusters to derive a directional vector $\hat{v}^*$. This is achieved by fitting a 3D line model through each cluster by using a RANSAC-based estimation. 
In the following step, an iterative search for the next lane marking center $p\subs{c}$ is done, based on the estimated directional vectors.
As the directional vectors $\hat{v}^*$ can contain high variations and inaccuracies, originating from partial scans or occlusion, further directional vectors $\hat{v}$ are stabilized through matched marking directions.
If a marking center was already matched, the directional vectors $\hat{v}$ are calculated with $\gamma = 0.5$ as:
\begin{equation}
\hat{v}\subs{i+1} = \gamma \cdot \hat{v}^*\subs{i} + (1-\gamma) \cdot \hat{v}^*\subs{i-1}
\label{eq:dir-vec-stabilization}
\end{equation}

Based on the directional vector, the next expected lane marking center $p\subs{c}$ is calculated through $p\subs{c} = \hat{v} \cdot d$. To account for offsets in the point cloud, on each $p\subs{c}$ a ball radius search for the next marking center is done.
Based on evaluations of different data sets, we define the maximum search distance $d\subs{\text{max}}$ for $d$ as 1.5 times the regulatory expected distance between lane marking centers. To compensate for different scanning mistakes and damaged markings, $p\subs{c}$ is calculated iteratively by raising $d$ in three-meter steps till $d\subs{\text{max}}$. At each $p\subs{c}$ the ball radius search is executed with half the regulatory lane width. If a marking center is found, it is set as a new search point and connected to the current center. The search continues as long as the next marking is found. 

The resulting connections between lane markings create an implicit model that partially or fully reassembles the overall lane marking stripe. 

To reduce wrong connections and higher stability against high reflectivity outliers in the point cloud, all steps utilize full 3D point cloud information, as described in \cite{eisemann-opendrive}.

\subsection{Refinement of lane markings}\label{subsec:comb-candid-lines}
To achieve a better analysis of the extracted and connected lane markings, further refinement is done similar to the previous Sec.~\ref{sec:gen-candid-lines}. 
By only taking into account starting and endpoints of connected markings, higher thresholds can be used to reduce partial connections of continuous marking stripes. This further reduces the amount of partial connections, especially when the ego vehicle changes lanes multiple times within the segment.

To ensure correct ordering within each connected line, markings are first sorted based on their distance to the segment origin. On this basis, another iterative search is started leveraging the method described in Sec.~\ref{sec:gen-candid-lines}. Using the last two points of each connection, the search is executed for  $d\subs{\text{max}}$ set as 3.5 times the regulatory expected distance between lane marking centers.
If a fitting starting point of a connection is found, it is assumed to be part of the same lane marking stripe and both connections are fused.

\subsection{Classification of Lane Markings}\label{subsec:classification-lane-markings}
Since the correct classification of the respective lane marking types is a key part of our overall process, we explain it in further detail. 
The aim of the classification is the ability to match the detected lane markings with the number of lanes provided by OSM in combination with regulatory definitions, such as solid lanes on the outer edges of the road. Therefore, our process currently focuses on the distinction between solid, dashed, and unknown markings.

Classification is done on two different measurements, the size of the lane markings and the distance between them. Based on the country and road type retrieved from the OSM data, the lane marking specifications are adjusted.

Through the analysis of the cluster size in Sec.~\ref{sec:gen-candid-lines}, solid lanes are detected with high confidence. This is due to the fact, that outliers tend to be only partial scans of markings, which would result in shorter clusters than the defined threshold. On the other hand, these outliers can lead to non-detection of solid lanes. 

As a straightforward distinction between solid and dashed lane markings, the size of the cluster is taken first into account.
To identify dashed markings, the size of more than 50\% of the lane markings connected must be in between $\pm$~10\% of the regulatory size. 
Evaluations of our data show, that in the case of partial scans of solid lane markings, the sizes of the clusters vary widely and the condition fails.

For lane marking stripes, where the previous classification was not able to distinguish between dashed or solid, the distance between the markings is evaluated.

Due to the distance-based resolution of LiDAR sensors, markings farther from the sensor might be undersampled and the resulting cluster size is underestimated, but the cluster's overall position remains accurate.
To exploit these phenomena, the average distance between markings is analyzed when more than four are available. If the average distance between marking centers is within $\pm$ 10\% of the legal requirements, the respective stripe is considered dashed. When the average distance is shorter than a single line marks' required length, the markings are classified as solid.

Cases that could not be matched or didn't meet the required minimum number of markings are defined as class unknown. Although their type is unknown, the extracted markings are used in further steps after matching the extracted lanes with OSM information.

\subsection{Calculation of Reference Line}\label{subsec:calc-ref-line}
\begin{figure}[tb]
\centering
\vspace{0.02in}
\includegraphics[width=0.995\columnwidth]{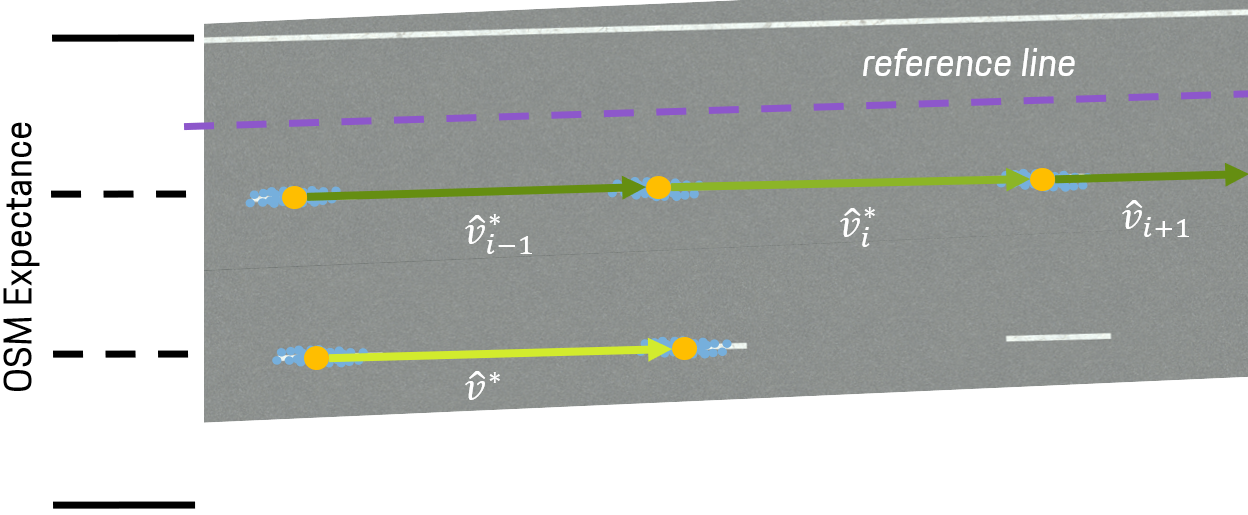}
\caption{Overview over lane generation and positional calculation. Lane marking point clouds are displayed in \textit{blue}, and respective estimated centers in \textit{orange}. The directional vectors are shown in shades of \textit{green}. Based on the expected number of lanes from OpenStreetMap, the position of the resulting \textit{reference line}, displayed in \textit{purple}, is calculated.}
\label{fig:method-overview-lanegen}
\end{figure}

The primary factor for the quality of an OpenDRIVE file is the accuracy of the reference line\cite{schwab-validation-2022}.

Position and calculation of the reference line in the geometry depends on the road type and country, which can be adjusted based on the OSM information described in Sec.~\ref{subsec:data-and-processing}. Since this enables the use of various road types, we focus on German highways for better comprehension. 

The proposed generation of the reference line is based on multiple steps. First, matching the extracted lane information with the information gathered from OSM. Second, estimating the offset of these lanes to the planned reference line position. Third, the calculation of the reference line, based on the extracted lane markings and estimated offset. Finally, the export of OpenDRIVE geometric elements.

For German highways, the position of the reference line is set to the middle of the left lane. This is because left-side highway exits are a rare case in Germany and therefore enable a unified handling of entry and exit lane sections.

The fusion of the known number of lanes, the type of road, the regulatory requirements for lane markings, and the width of the lane, allows the estimation of the offset of the extracted lane markings to the reference line. 
In the example case of a German highway, shown in Fig.~\ref{fig:method-overview-lanegen}, it is known from OSM that this segment consists of three lanes. Further regulations specify that highway road boundaries are required to be marked with solid markings. If two parallel dashed lane markings were extracted from the LiDAR data, it can be assumed that these mark the boundaries of the middle lane. Based on this conclusion and the known lane width, the offset to the reference line can be calculated.

While this example is trivial, complexity increases with a higher number of lanes and the existence of emergency shoulder lanes. Moreover, the solution relies on the lanes extracted and classified from the real data, which is why we base the subsequent explanation on the detection results.

In the simplest case, one or more solid lanes are extracted in the segment. As regulations stipulate solid marking on the road boundaries, these can be matched between left and right boundary depending on their position relative to the segment origin.
If only dashed lane markings are detected, they can be matched similarly, as long as the number of detected markings equals the number of expected dashed markings. 

Since with a higher number of lanes, the result can be ambiguous, we utilize the relative position of the extracted lanes in the point cloud. To achieve a more accurate match, the point cloud generated in Sec.~\ref{subsec:line-marker-extrac} is cropped based on the extracted lane markings. For each lane marking, only points are kept with a distance lower than the maximum distance to the road boundary. The union of these points is used to match the lanes based on the estimated distance towards the road boundary.
This methodology is also used in the case where no lane could be classified.

The generation of the reference line makes use of all extracted lane markings. For each lane marking the normal vector towards the targeted reference line position is calculated. Based on the previously assigned offset, a point on the normal vector is created. This results in a high-resolution reference line, including all available marking information.

\subsection{Export of OpenDRIVE}\label{subsec:export-opendrive}
\begin{figure}[t]
\centering
\includegraphics[width=1\columnwidth]{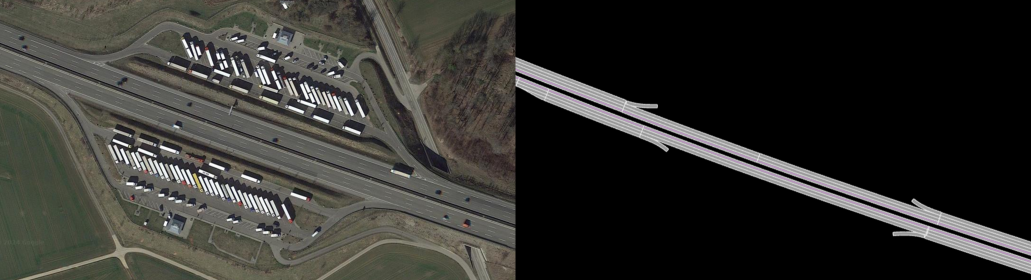}
\caption{Comparison of Google Maps satellite imagery against our generated OpenDRIVE example. The main highway is based on the calculation of the reference line, while the off- and on-ramp are derived from OpenStreetMap.}
\label{fig:ExitEntry-real-to-esmini-comparison}
\end{figure}

While the previous steps can be processed independently from one another, for the export of the OpenDRIVE consecutive segments are needed.
The individual segments are collected and sorted according to timestamps and OSM relation. Segments without a result from previous steps are collected separately for manual quality control and adjustment.

To obtain a continuous reference line, we reprocess it to avoid geometric leaps and kinks\cite{schwab-validation-2022}. 
Since our processing is based on individual lane markings, the segment split described in \ref{subsec:segment-generation} can fall between dashed markings. Without correction, this would lead to OpenDRIVE segments with gaps equal to the space between line markings. While correct on a measurement level, the OpenDRIVE standard stipulates a continuous description without gaps. To fulfill this requirement, the starting point of the successor element is added as the endpoint of the current segment reference line. 
This process is repeated on all elements and further enables automated quality control. If the lateral offset between reference lines exceeds a certain threshold these segments are flagged for manual review.

Exporting a single OpenDRIVE road element per generated segment leads to a high number of roads with equal information. To reduce this number, we follow the OSM structure and create a single OpenDRIVE road per OSM \textit{way}, with segments describing the geometry of the road.

For the generation of complex geometries from measurement data, the OpenDRIVE standard recommends the use of parametric cubic curves\cite{asam-opendrive}.
As described in Sec.~\ref{sec:mapformat}, it is important to prevent geometric leaps and kinks in the final OpenDRIVE map. 
Following \cite{eisemann-opendrive}, we introduce an importance weighting for fitting the parametric curves through the raw reference line. As the start and end points between the segments have been aligned in the previous step, we assign higher weights to these, resulting in a closer fit of the curve and therefore a reduction of geometric leaps.

Road elevation and super-elevation are added as additional cubic curves along the fitted reference line. These are also calculated for each segment separately directly from the extracted reference line markers, omitting the previously described weighting. 

For each OSM \textit{way}, an OpenDRIVE road element is generated, where the calculated cubic curves of the segments describe the geometry. Information about the size of the lanes and their type is added by combining the OSM information with regulatory information.
Through the OSM-derived order of the roads, the respective predecessor and successor attributes are set, as well as lane links are set up.

If the number of lanes changes from one element to another, two separate lane sections are set up, which fade in or fade out the changing lane.

Entry and exit lanes represent a special case. As their existence is known from OSM, an OpenDRIVE junction is generated. This contains the respective predecessor and successor roads extracted. To comprehensively model these junctions a placeholder road is created as a ramp. For this, the starting position of the element is estimated through the calculated reference line, the known road width, and the size of the ramp. On the approximated starting position of the ramp a simple new road object is created, described by an OpenDRIVE spiral geometry. The ramp is added to the junction and lanes are assigned the correct types and links.

\section{Result and Discussion}\label{sec:results-discussion}
In this section, we compare the results of the generated OpenDRIVE files quantitatively against other maps and evaluate their quality by simulating an original driving scenario.

\subsection{Quantitative Analysis}
\begin{table}[t]
\caption{For accuracy we compare against PEGASUS \cite{pegasus-project} HD maps and for reproducibility against our methodology.}
\centering
\begin{tabular}{@{}lcc@{}}
\toprule
                & \multicolumn{1}{l}{agst. PEGASUS HD} & \multicolumn{1}{l}{agst. self (ours)} \\ \midrule
RMSE            & 0.337 m                                 & 0.274 m                                   \\ \midrule
avg. distance   & 0.243 m                                 & 0.213 m                                   \\ \midrule
std. deviation $\sigma$ & 0.201 m                                 & 0.166 m                                   \\ \midrule
eval. length    & 44.8 km                                 & 30.6 km                                   \\\bottomrule
\end{tabular}

\label{tab:quant-eval}
\end{table}

\begin{figure*}[!tp]
	\centering
	\subfloat[1][\label{fig:qual-examples-1}]
	{\includegraphics[width=0.68\columnwidth]{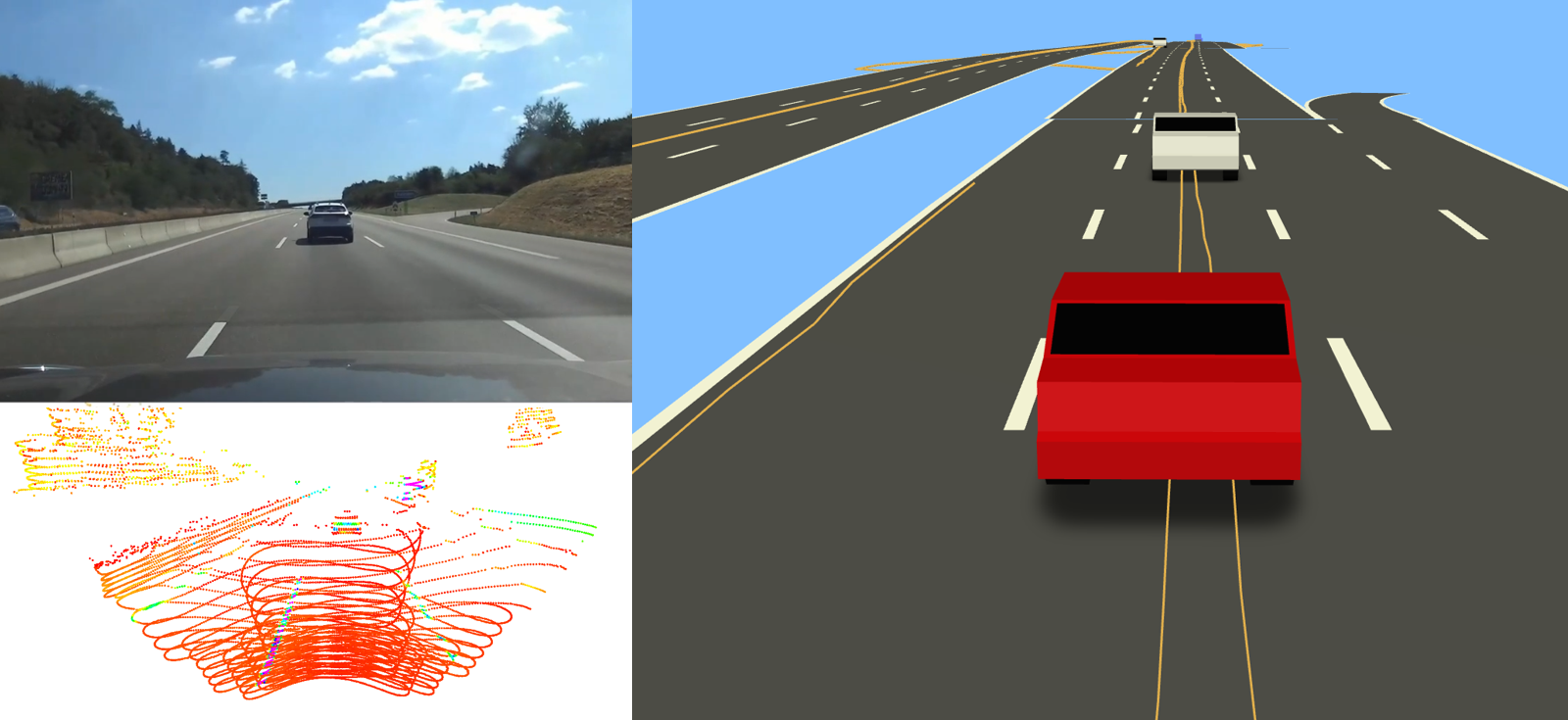}}\hfill
	\subfloat[1][\label{fig:qual-examples-2}]
	{\includegraphics[width=0.68\columnwidth]{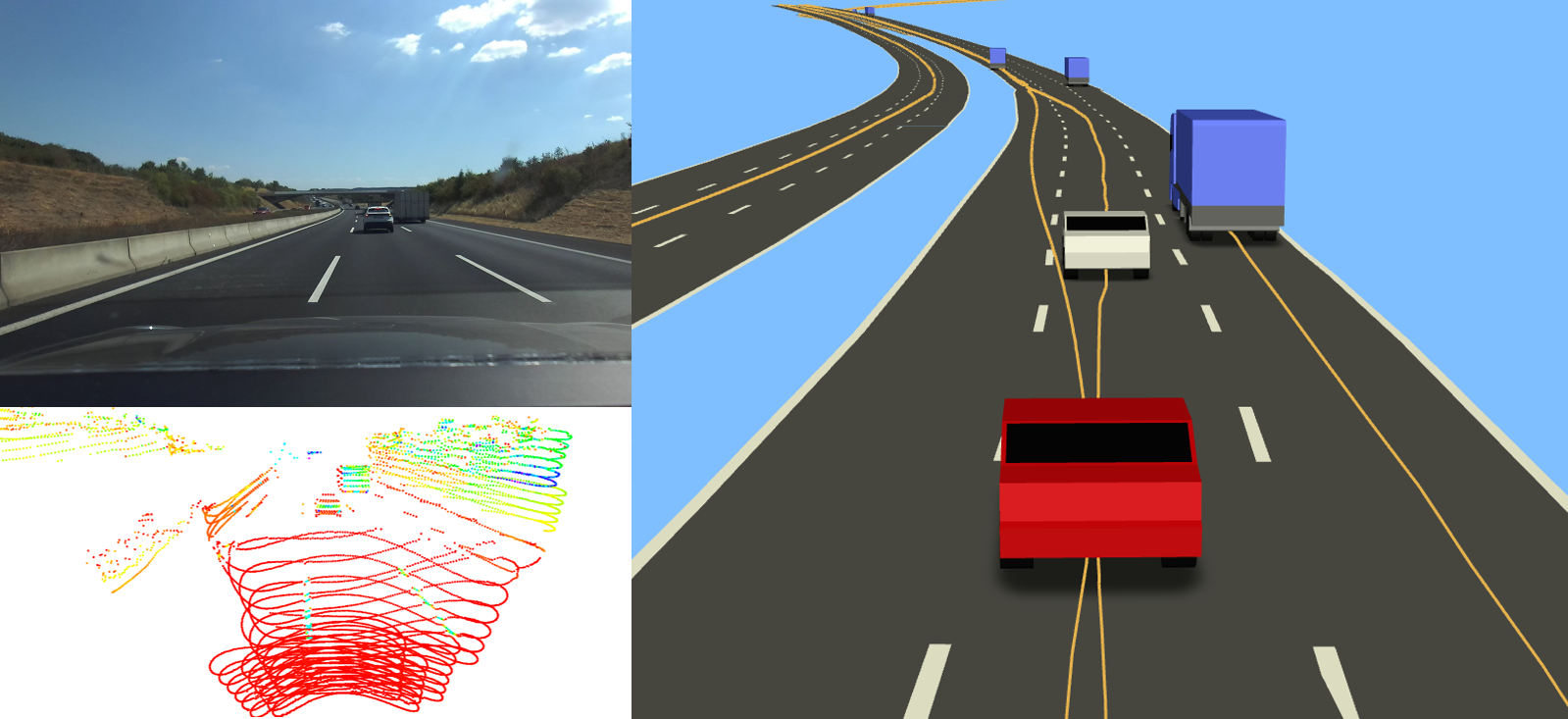}}\hfill
	\subfloat[1][\label{fig:qual-examples-3}]
	{\includegraphics[width=0.68\columnwidth]{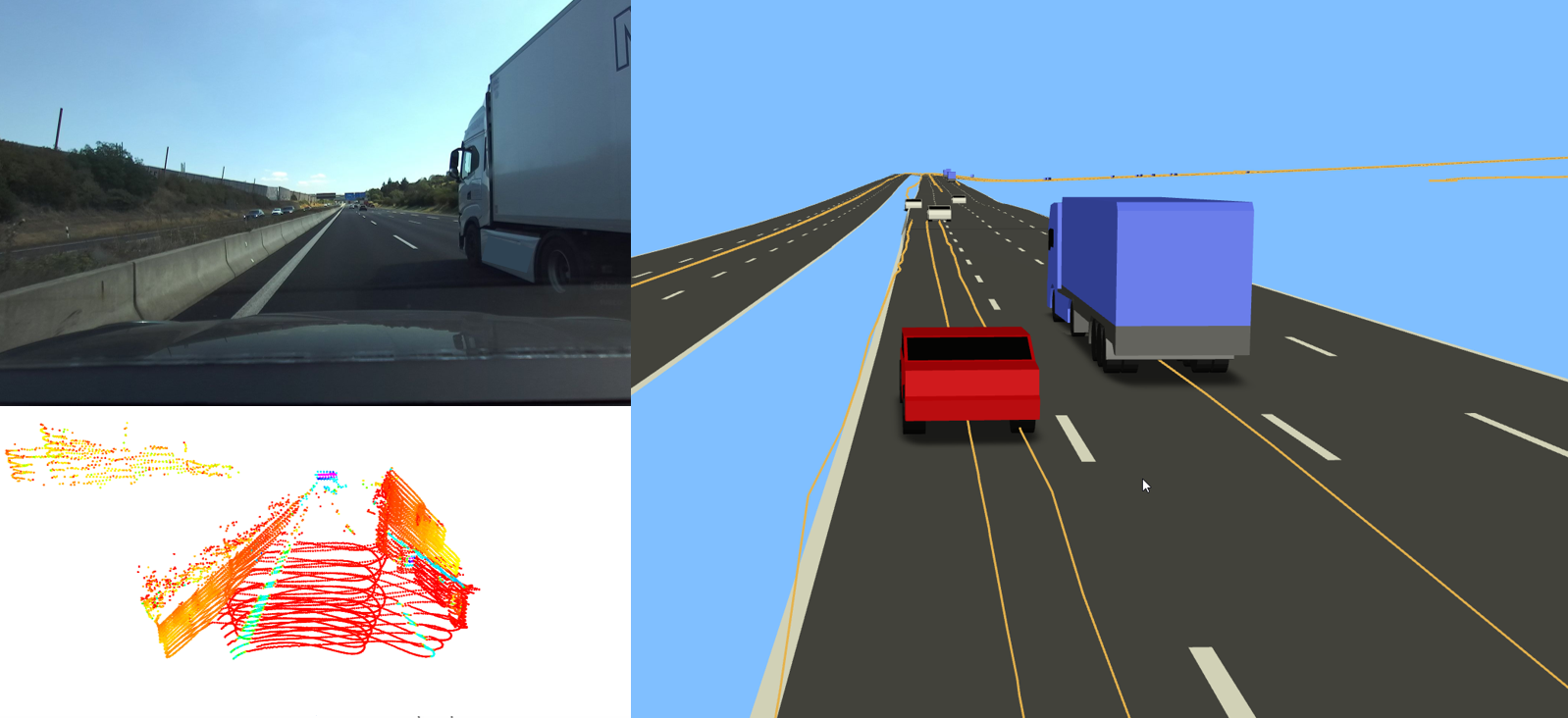}}\hfill
	\caption{Excerpts of one drive for qualitative comparison. The respective camera frame and LiDAR scan from the original recording are referenced in each image. The resulting OpenDRIVE is displayed as esmini \cite{esmini-environment-2023} rendering, including traffic participants from generated OpenSCENARIO\cite{asam-openscenario}.}
	\label{fig:qualitative-example-images}
 
\end{figure*}

For a comprehensive quantitative evaluation, we examine three aspects. 
As our main focus is to enable industrial-scale road generation, we present an overview of the success rate and the processing time. 
To evaluate the accuracy, we compare our generated OpenDRIVE files with publicly available HD OpenDRIVE maps, and subsequently, analyze the reproducibility of the accuracy.

Across 286 km of processed data, our proposed method successfully extracts reference line information and converts it into OpenDRIVE for 91.6\% of the total distance. Examination of the failed segments reveals, that most are due to erroneous scans of lane marks or segments lacking lane markings.
As segment processing can be parallelized on a large scale, we evaluate the processing time per segment. Including the map-matching and export into OpenDRIVE, the processing time per segment is approximately equal to the real-world driving time for that particular segment.

The OpenDRIVE standard provides a multitude of ways to describe road layouts, leading to equivalent roads with identical or near-identical geometry but different parameters\cite{eisemann-opendrive}. 
One reason is the possibility of offsetting lanes and markings along the reference line on multiple occasions. Although beneficial from a user perspective, this introduces different challenges for the comparison of road models. 
Therefore, recent research performs the comparison on the resulting geometry rather than of the parameters describing them.
Based on these aspects, we follow the evaluation proposed by \cite{chiang_automated_2022, eisemann-opendrive}. Similarly, the comparison is done by sampling uniformly spaced points of each generated road reference line and evaluating the Root Mean Square Error (RMSE), average distance, and standard deviation ($\sigma$). 

Since the related work presented in Sec.~\ref{subsec:related-work} relies on vehicle or sensor characteristics, a direct comparison is not possible. To achieve a reliable accuracy metric, we therefore compare against HD maps from the PEGASUS project\cite{pegasus-project}.

As our proposal leverages the feature extraction and lane generation of \cite{eisemann-opendrive}, we provide evidence of similar accuracy.
The input data consists of the same drive recordings as in \cite{eisemann-opendrive}. These consist of four different recordings, taken over a period of 8 weeks, following the same part of the German highway A8. The overall distance on the part accumulates to a total of 44.8 km, including 31 cut-in scenarios and 29 ego vehicle lane changes. The ego vehicle's speed varies between 80 and 140 km/h, averaging 116 km/h, and the weather conditions range from sunny to mildly rainy.

The results of the comparison between the generated 44.8 km of OpenDRIVE files against the PEGASUS OpenDRIVE files are shown in Tab.\ref{tab:quant-eval}. Our method achieves an average nearest neighbor distance of 0.243 m, an $\sigma$ of 0.201 m, and an RMSE of 0.337 m. In accordance with \cite{eisemann-opendrive}, our results fulfill the requirements of the Taiwan HD map standards \cite{taiwan-hd-map-standard-2020}, making our result an HD map as well. Further, the results show that our approach incorporates the accuracy of the lane estimation while providing additional benefits for large-scale adaption. Moreover, our results demonstrate that our proposed reference line position estimation and calculation provide stable results.

For an industrial use case, the reproducibility of the results is also an important factor. For the evaluation, one of the recording drives is chosen randomly as ground truth, which the remaining three are compared against. This yields an average distance of 0.213 m, with an $\sigma$ of 0.166 m and RMSE of 0.274 m. The observed offsets hereby originate from different factors along the methodology. As initially mentioned, we only introduce moderate requirements on the calibration and offsets in the recorded data.  Consequently, map generation is influenced by calibration errors, GNSS inconsistencies, the alignment of geo-references, and offsets in the geometry fitting. Despite these offsets, the findings demonstrate that our approach consistently produces reliable maps, across diverse traffic scenarios and driving conditions.
\subsection{Qualitative Analysis}
To test the consistency of the generated OpenDRIVE in simulating a real-world driving scenario, we generate an additional OpenSCENARIO \cite{asam-openscenario} file. Information about traffic participants is hereby obtained from the series sensors of the ego vehicle. To obtain the needed unified coordinate system, these are transformed from the vehicle coordinate system into the global through leveraging the same transverse Mercator projection as described in Sec.~\ref{subsec:line-marker-extrac}.

Excerpts from the corresponding esmini \cite{esmini-environment-2023} rendering of the generated files are displayed in Fig.~\ref{fig:qualitative-example-images}. The full video of the drive comparison is available on our data exchange\footnote{\url{https://dataexchange.porsche-engineering.de/wl/?id=Duoz69wGMqdyo2yGGsgyTJTXalxGTLzb} \\Password: IEEE-IV24}.
The presented visualization demonstrates the usability of our approach for the derivation of driving scenarios. Even in complex situations with a high number of vehicles, e.g. in Fig.~\ref{fig:real-to-esmini-comparison}, traffic participants are accurately placed on each lane and within each lane. Observed maneuvers, like cut-in and cut-out, are replicated correctly in the simulation.  Fig.~\ref{fig:qual-examples-1},\ref{fig:qual-examples-2} highlight the lateral accuracy of our approach. The first shows the target vehicle on the far right of the respective lane, consistent with the input data. Likewise, the second example shows similar for the ego vehicle lane change.

This is enabled through the close-to-real shape of the generated road representation. Despite the LiDAR FoV being occluded to a high degree, our approach is able to reconstruct a realistic road representation, as for example in Fig.~\ref{fig:qual-examples-3}. Also, the examples demonstrate the effectiveness of the OSM data integration, as entry and exit lanes are placed correctly on the segment. This allows a better understanding of lane change behavior in future work.

Given that all vehicles in the simulation are positioned relative to the GNSS with an accuracy of $\pm$5 cm we conclude that the road has a high degree of realism. Even in situations where the GNSS has a higher offset, e.g., through loss of GNSS signal, our method is capable of accurately inferring the road structure and displays a smooth and homogeneous transition between geometric elements.

\section{Conclusion and Future Work}
Through the importance of road networks for highly automated driving functions and their evaluation, research on the creation of such has seen considerable interest over the last few years. In this paper, we proposed an approach for the generation of high-resolution road representation from real-world test drives. In contrast to previous research, our focus lies not only on the accuracy of the method but more on the scalability for multiple hundred kilometers. Leveraging sparse information from OpenStreetMap, the original test drive is divided into road segments. For each segment, the sparse LiDAR data is accumulated and through the extraction of lane markings, the reference line is estimated. To achieve the final OpenDRIVE format, the geometry of the individual segment is created, together with entry/exit lanes, and the relations, such as junctions, are set.
The segment-wise approach hereby offers greater flexibility than previous methods. Besides the scalable parallel processing of the segments, these can further be modified, corrected, and stored. Even in cases where the processing for a segment does not succeed, only the respective segment is affected.

Furthermore, we could show that the algorithm's results fulfill the requirements for HD map standards and driving function simulation, which could be shown through quantitative and qualitative metrics. The resimulation of the input test drives shows the accurate placement of traffic participants close to reality, even at ego lane changes.

\bibliographystyle{IEEEtran}
\bibliography{bibopendrive}

\end{document}